\documentclass[journal]{IEEEtran}
\usepackage{amsmath}
\usepackage{amssymb}
\usepackage{mathtools}
\usepackage{multirow}
\usepackage{cite}
\usepackage{array}
\usepackage{url}
\usepackage{graphicx}
\usepackage{booktabs}
\usepackage[english]{babel}
\usepackage[utf8]{inputenc}

\begin{document}

\title{Learning Biomechanically Plausible Human\\ Motion from Sparse Radar Point Clouds}

\author{Jonas Leo Mueller$^{1,2,3}$,
        Markus Gambietz$^{4}$,
        Alexander Weiss$^{1,4}$,
        Daniel Krauss$^{1,3}$,
        and Bjoern M. Eskofier$^{1,2,3,5}$\\[0.6em]
{\small $^{1}$Department Artificial Intelligence in Biomedical Engineering (AIBE), Friedrich-Alexander-Universit\"at Erlangen-N\"urnberg, Erlangen, Germany}\\
{\small $^{2}$Chair of AI-supported Therapy Decisions, Ludwig-Maximilians-Universit\"at M\"unchen, Munich, Germany}\\
{\small $^{3}$Munich Center for Machine Learning (MCML), Munich, Germany}\\
{\small $^{4}$Chair of Autonomous Systems and Mechatronics, Friedrich-Alexander-Universit\"at Erlangen-N\"urnberg, Erlangen, Germany}\\
{\small $^{5}$Institute of AI for Health, Helmholtz Zentrum M\"unchen, Neuherberg, Germany}\\[0.4em]
{\small Corresponding author: Jonas Leo Mueller (jonas.leo.mueller@fau.de)}%
\thanks{This work was partly funded by the Deutsche Forschungsgemeinschaft (DFG, German Research Foundation) -- SFB 1483 -- Project-ID 442419336, EmpkinS. This work has been submitted to the IEEE for possible publication. Copyright may be transferred without notice, after which this version may no longer be accessible.}}

\maketitle

\begin{abstract}
Radar-based human pose estimation has focused on improving learning algorithms while representing the body as unconstrained keypoint coordinates. We address the underexplored dimension of anatomical fidelity by integrating a full-body skeletal model into a differentiable, end-to-end trainable radar-based pose estimation framework, in which the pose network is supervised through forward kinematics while subject-specific geometry is fitted beforehand. Subject-specific body segment proportions are predicted from radar point cloud features to scale a biomechanical skeleton. A motion prediction network maps temporal radar sequences to generalized coordinates, and differentiable forward kinematics converts predicted joint angles into 3D positions. A contact classification loss encourages physically plausible foot-ground interaction. Under leave-one-subject-out cross-validation on 11 healthy participants performing rehabilitation exercises, the framework achieves $6.456 \pm 1.759$\,cm mean per-joint position error (MPJPE), $8.083 \pm 0.884^{\circ}$ mean per-joint angle error (MPJAE), $0.935 \pm 0.009$ contact classification F1, and $3.4 \pm 1.3\,\%$ scaling error. This proof-of-concept study demonstrates the feasibility of recovering interpretable biomechanical descriptors from a single low-cost radar sensor in a controlled laboratory setting, a prerequisite for future clinical motion analysis.
\end{abstract}

\begin{IEEEkeywords}
biomechanical modeling, forward kinematics, ground contact classification, human pose estimation, millimeter-wave radar
\end{IEEEkeywords}

\noindent\textbf{\textit{Impact Statement---}}A single radar sensor combined with a differentiable skeletal model can recover clinically interpretable joint angles and ground contact from sparse point clouds, a step toward unobtrusive biomechanical monitoring without cameras or body-worn sensors.

%% ====================================================================
\section{INTRODUCTION}
%% ====================================================================

\IEEEPARstart{U}{nobtrusive} monitoring of human movement is critical for early detection and treatment of movement-related conditions, from neurological disorders such as Parkinson's disease \cite{wang2020early} to musculoskeletal conditions including rheumatoid arthritis \cite{visser2005early}. Camera-based systems inherently record identifiable patient imagery, raising privacy concerns and regulatory requirements under the EU Artificial Intelligence Act \cite{act2024eu}. Continuous monitoring therefore requires sensors that preserve privacy and do not interfere with natural movement \cite{krauss2024review, hasty1995early}. Frequency-modulated continuous-wave (FMCW) millimeter-wave radar avoids these concerns by measuring radio-frequency reflections rather than recording visual imagery, while operating independently of ambient lighting \cite{engel2025advanced, ho2024rt, lee2023hupr, mueller2024end, mueller2025radproposer, richards2005fundamentals, krauss2024review}. Radar-based human pose estimation (HPE) methods predict 3D keypoints from either constant false alarm rate (CFAR) detected point clouds \cite{engel2025advanced, sengupta2020mm, sengupta2020nlp, sengupta2022mmpose, hu_mmpose-fk_2024, salehin2025radar, an2022, an2022fast, su2025posegraphnet, chiang2024enhancing, luo2025} or radar tensor representations \cite{mueller2024end, mueller2025radproposer, lee2023hupr, ho2024rt, cao2023task, cao2022joint}. We operate on point clouds, since full radar tensors incur memory and computational costs prohibitive for real-time rehabilitation settings \cite{mueller2025adaptive}.

While these methods have achieved substantial progress in learning representations from radar signals, one fundamental aspect remains underexplored, namely the anatomical fidelity of the underlying human model. All existing radar-based HPE approaches, including the widely adopted mmMesh baseline~\cite{xue2021mmmesh}, represent the body as a set of unconstrained 3D keypoint coordinates \cite{engel2025advanced, sengupta2020mm, sengupta2020nlp, sengupta2022mmpose, hu_mmpose-fk_2024, salehin2025radar, an2022, an2022fast, su2025posegraphnet, chiang2024enhancing, luo2025, lee2023hupr, ho2024rt, cao2023task, cao2022joint}, leaving the network free to produce fluctuating bone lengths and biomechanically implausible postures. Moreover, without awareness of the ground plane, such models cannot enforce plausible foot-ground interaction, which results in feet that slide along or penetrate the floor. Some methods introduce limited anatomical constraints. Hu et al.\ \cite{hu_mmpose-fk_2024} add a forward kinematics layer that predicts per-frame bone lengths and Euler angle rotations along a simple kinematic chain, but their representation still permits bone length fluctuation across frames and does not enforce anatomically defined joint axes or consistent subject-specific scaling. Salehin et al.\ \cite{salehin2025radar} propose cycle consistency through micro-Doppler supervision. However, no existing method integrates a full-body skeletal model with physiologically grounded joint definitions that constrains predictions to kinematically consistent configurations by design.

In biomechanical analysis, the human body is represented as a kinematic tree of rigid segments connected through joints with anatomically defined degrees of freedom \cite{delp2007opensim, rajagopal2016full}. Each segment corresponds to a physical bone, joint rotation axes are aligned with physiological axes derived from the biomechanical literature, and segment dimensions are scaled to individual anthropometry. The model state is parameterized by generalized coordinates, including joint angles. Predicting joint angles rather than unconstrained Cartesian coordinates means that forward kinematics analytically maps the output to 3D positions, guaranteeing consistent limb proportions, kinematic chain dependencies, and subject-specific morphology by construction. Because the generalized coordinates parameterize a low-dimensional manifold of kinematically consistent poses, the network explores a vastly reduced and structured output space compared to unconstrained Cartesian regression. Moreover, the predicted joint angles are directly interpretable by clinicians without any post-hoc conversion, as they are the native quantities of gait analysis and musculoskeletal assessment. Combined with a contact classification objective, the model becomes explicitly aware of the ground plane, encouraging zero foot velocity and correct height during stance. However, fitting such models has traditionally required laboratory-grade instrumentation such as marker-based optical motion capture \cite{uhlrich2023opencap, gozlan_opencapbench_2025} or inertial measurement unit arrays \cite{dorschky2019estimation}, which is impractical for continuous home monitoring. Markerless vision-based approaches \cite{uhlrich2023opencap} reduce instrumentation burden but still depend on cameras with inherent privacy concerns. Despite the proven value of biomechanical analysis in rehabilitation and sports science \cite{roupa2022modeling, smith2021review, Grabiner2008TrunkKA, Martino2020NeuromechanicalAO}, no radar-based HPE method has incorporated a full-body skeletal model with subject-specific scaling and ground contact classification.

In this work, we present a proof-of-concept framework that recovers kinematically consistent full-body motion directly from radar point clouds using a single FMCW sensor. By combining a differentiable skeletal model with radar-based subject-specific scaling and contact classification losses, our approach produces interpretable kinematics and ground contact estimates without body-worn sensors or multi-camera setups. Evaluated on healthy participants performing rehabilitation exercises, this work demonstrates the feasibility of bringing clinically interpretable biomechanical descriptors to a low-cost, privacy-preserving sensor. The main contributions are as follows:
\begin{enumerate}
\item The first integration of a full-body skeletal model into radar-based human pose estimation, achieving a mean per-joint position error of $6.456 \pm 1.759$\,cm and an MPJAE of $8.083 \pm 0.884^{\circ}$ on rehabilitation exercises.
\item Subject-specific scaling of the skeletal model directly from radar point cloud features, achieving a scale prediction error of $3.4 \pm 1.3$\,\%.
\item A ground contact classification objective that encourages physically plausible foot-ground interaction, achieving an F1 score of $0.935 \pm 0.009$.
\end{enumerate}

\begin{figure*}[t]
\centering
\includegraphics[width=\textwidth]{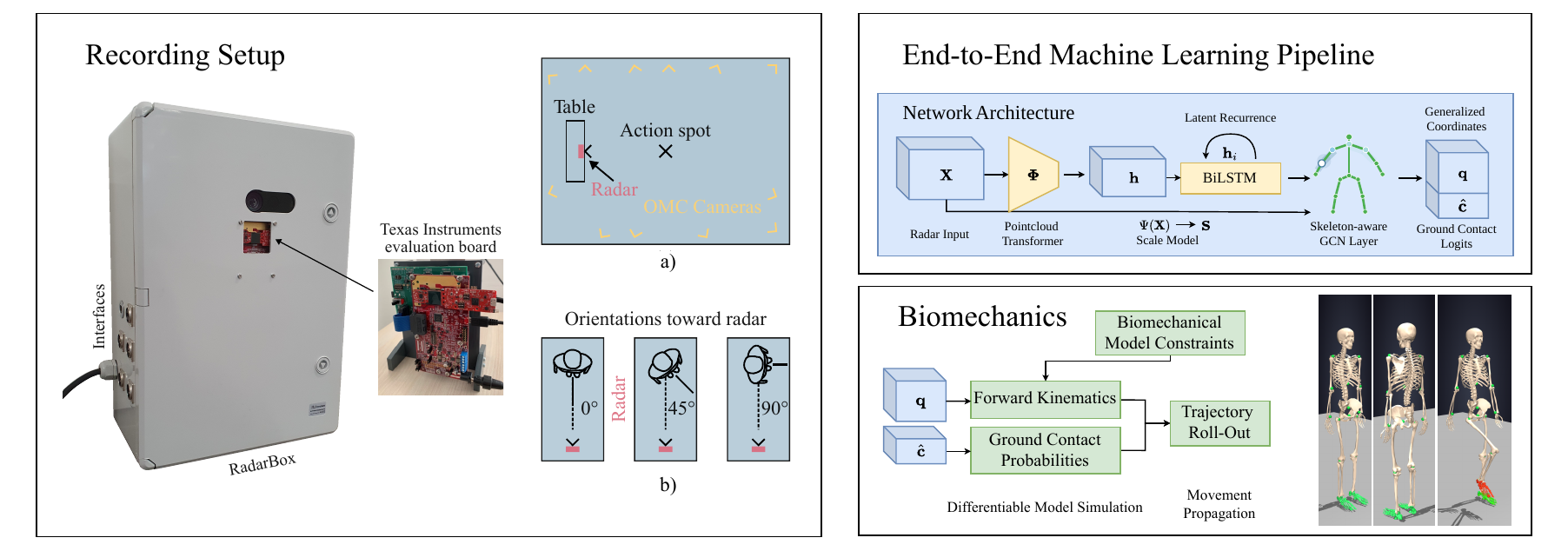}
\caption{Overview of the proposed framework. Left: Recording setup showing the RadarBox housing a Texas Instruments FMCW evaluation board, the laboratory layout with radar and optical motion capture (OMC) cameras (a), and (b) the three aspect angles ($0^\circ$, $45^\circ$, $90^\circ$) used to evaluate robustness to participant orientation. Right top: End-to-end network architecture where radar point clouds are transformed into latent features, a BiLSTM with latent recurrence predicts generalized coordinates and ground contact logits, and a scale model maps point cloud features to subject-specific body segment proportions. Right bottom: Differentiable biomechanical simulation where predicted joint angles and contact probabilities are processed through forward kinematics with anatomical constraints, producing a temporal trajectory roll-out of the biomechanical skeleton.}
\label{fig:graphical_abstract}
\end{figure*}

%% ====================================================================
\section{MATERIALS AND METHODS}
%% ====================================================================

We first describe the dataset (Sec.~\ref{sec:dataset}). The pipeline is built around a differentiable biomechanical skeleton (Sec.~\ref{sec:biomech}). Virtual markers map ground-truth joint centers onto this skeleton to enable supervision (Sec.~\ref{sec:marker_placement}), and subject-specific body proportions predicted from radar scale the skeleton template (Sec.~\ref{sec:scaling}). A motion prediction network maps temporal radar sequences to generalized coordinates and ground contact probabilities (Sec.~\ref{sec:motion_net}), and differentiable forward kinematics converts the predicted joint angles into 3D marker positions (Sec.~\ref{sec:fk}). A combined data-tracking and contact classification loss drives training (Sec.~\ref{sec:contact_training}). At inference, the entire pipeline operates solely from radar point clouds without requiring motion capture or body-worn sensors.

\subsection{Dataset}
\label{sec:dataset}

We use the openly available mmRadPose dataset \cite{engel2025advanced}\footnote{\url{https://zenodo.org/records/14738837}}, which provides synchronized millimeter-wave radar and optical motion capture recordings. The dataset contains 11 participants performing 11 rehabilitation exercises (left, right, and bilateral upper limb extension, biceps curls, front arm rotation, torso forward bending, left and right front lunge, squats, side lower limb extension, and front lower limb extension), each recorded at three aspect angles ($0^\circ$, $45^\circ$, and $90^\circ$) with respect to the radar boresight. Radar data is captured with a 60\,GHz FMCW radar (see Appendix, Table~\ref{tab:radar_params}). Each raw CFAR-detected frame contains up to 64 points; we aggregate three consecutive frames into superframes of up to $N = 192$ points, where each point carries 3D spatial coordinates, reflection intensity, and 3D Doppler velocity as features ($F = 7$). For full details on the recording setup, we refer the reader to \cite{engel2025advanced}.

Ground truth consists of 26 3D joint center positions (keypoints) from an optical motion capture system synchronized via hardware trigger (see Appendix, Fig.~\ref{fig:gt_skeleton}), of which we use a subset of $M = 17$ that corresponds to body segments on the biomechanical model (Sec.~\ref{sec:biomech}).

\subsection{Biomechanical Model}
\label{sec:biomech}

We adopt a full-body skeletal model based on the kinematic structure of the Rajagopal model \cite{rajagopal2016full}, implemented in a fully differentiable framework using PyTorch. The model comprises 20 rigid bodies organized as a kinematic tree rooted at the pelvis. It is parameterized by $N_q = 37$ generalized coordinates $\mathbf{q} \in \mathbb{R}^{37}$, consisting of a six-degree-of-freedom free joint for the pelvis root and 31 hinge joints for anatomical articulations, where multi-degree-of-freedom joints such as the hips, shoulders, and lumbar spine are each represented by multiple co-located hinge axes with anatomically defined rotation directions. The kinematic tree topology and the rotation axes of all hinge joints are derived from the human biomechanical literature \cite{rajagopal2016full}. The model defines 4 foot bodies (bilateral calcaneus and toes) for ground contact classification and supports $M = 17$ virtual markers for pose supervision (see Appendix, Table~\ref{tab:markers}). The parameterization enforces fixed segment lengths, a valid kinematic chain, and anatomically defined rotation axes by construction; it does not impose joint range-of-motion limits or self-collision constraints, so individual predicted configurations may still lie outside the physiological workspace.

\subsection{Marker Placement}
\label{sec:marker_placement}

Because the motion capture and biomechanical skeletons define different kinematic structures, we map the 17 selected ground-truth joint centers to virtual markers on the biomechanical model, each parameterized by a body-local offset. Since these offsets vary with individual morphology, we learn a mapping from subject-specific scale factors to body-local marker offsets. To obtain ground-truth offsets, we align each subject's motion-capture T-pose skeleton to the biomechanical model via rigid Procrustes alignment. Each marker's offset is then the displacement from its parent body origin, rotated into the body-local frame:
\begin{equation}
\mathbf{d}_j = \mathbf{R}_i^{\top} \left(\mathbf{m}_j^{} - \mathbf{p}_i\right),
\label{eq:marker_offset}
\end{equation}
where $\mathbf{m}_j^{}$ is the aligned marker position, and $\mathbf{p}_i$ and $\mathbf{R}_i$ are the position and orientation of its parent body from forward kinematics at the T-pose. A Ridge regression model, fitted to motion capture ground truth during training, maps the 20 scale factors $\mathbf{s}$ to the full set of $M \times 3 = 51$ body-local offsets $\mathbf{d} \in \mathbb{R}^{M \times 3}$, predicting where each marker sits in its parent body's local frame given the subject's proportions. Because both the scaling regression (described next) and the marker offset regression are trained from motion capture labels, the learned mappings generalize to inference where only radar observations are available, enabling a fully personalized skeletal model without motion capture instrumentation.

\subsection{Subject-Specific Scaling}
\label{sec:scaling}

Each body segment is scaled by a subject-specific factor to account for individual proportions. Our scaling approach predicts these factors directly from radar point cloud sequences.

We extract 19 geometric features per radar frame capturing spatial distribution, reflectivity statistics, and point density (see Appendix~\ref{sec:scale_prediction} for details). Temporal aggregation of mean and standard deviation yields a 38-dimensional feature vector per participant. A multi-task Lasso model maps this 38-dimensional input to 10 independent scale residuals, which are added to a default scale of 1.0 and mirrored bilaterally to produce a full 20-body scale vector $\mathbf{s} \in \mathbb{R}^{20}$.

Ground-truth scale factors for training supervision are computed from optical motion capture data. For each body $i$, the scale factor is the ratio of the measured segment length to the default model segment length:
\begin{equation}
s_i = \frac{\left\| \mathbf{j}_{c(i)} - \mathbf{j}_{p(i)} \right\|}{\left\| \bar{\mathbf{p}}_i - \bar{\mathbf{p}}_{p(i)} \right\|},
\label{eq:scale}
\end{equation}
where $\mathbf{j}_{c(i)}$ and $\mathbf{j}_{p(i)}$ are the ground-truth joint positions of the child and parent of body $i$, and $\bar{\mathbf{p}}_i$ denotes the world-frame body origin in the default model pose. Segments with direct joint-pair correspondences use per-segment ratios; segments spanning multiple model bodies share a uniform chain-based ratio. During forward kinematics, each body offset $\mathbf{o}_i$ is multiplied by its scale factor $s_i$, adapting joint positions to the individual morphology.

\subsection{Motion Prediction Network}
\label{sec:motion_net}

The motion prediction network is a sequence-to-sequence model that maps $T = 64$ superframes (each aggregating three consecutive CFAR-detected point clouds, Sec.~\ref{sec:dataset}) to $T$ pose predictions. For each input frame $\mathbf{X}_t \in \mathbb{R}^{N \times F}$, the network predicts generalized coordinates $\mathbf{q}_t \in \mathbb{R}^{N_q}$ and ground contact logits $\hat{\mathbf{c}}_t \in \mathbb{R}^{4}$, which are converted to probabilities via the sigmoid function during loss computation. The architecture consists of three stages; full architectural details are provided in Appendix~\ref{sec:arch_details}.

\subsubsection{Point cloud encoder}
Each aggregated radar frame is an unordered set of $N$ points with the $F = 7$ features described in Sec.~\ref{sec:dataset}. Since point clouds lack a canonical ordering, we use a transformer encoder without positional encodings to ensure permutation invariance. Per-point features are projected to a $d$-dimensional embedding, and a learnable CLS token is prepended. Self-attention aggregates information from all points into the CLS token, producing a global frame representation $\mathbf{h}_t \in \mathbb{R}^{d}$. All $T$ frames are encoded in parallel.

\subsubsection{Temporal model}
The sequence $\{\mathbf{h}_1, \ldots, \mathbf{h}_T\}$ is processed by a bidirectional LSTM with a residual connection and layer normalization, capturing past and future context within the temporal window to produce temporally contextualized features $\mathbf{z}_t \in \mathbb{R}^{d}$.

\subsubsection{Skeleton-aware GCN decoder}
A graph convolutional network (GCN) projects each $\mathbf{z}_t$ to per-body node features on the 20-node graph defined by the biomechanical model's kinematic tree (Sec.~\ref{sec:biomech}) and refines them through residual Chebyshev spectral graph convolution blocks \cite{defferrard2016convolutional}. Two output heads operate on the per-body node features. A joint angle head predicts the generalized coordinates $\mathbf{q}_t$ with bilateral weight sharing between left and right body pairs, and a contact head produces logits for the four foot bodies (calcaneus and toes on each side).

%\begin{figure}[t]
%\centering
%\includegraphics[width=\columnwidth]{fig/network_model.pdf}
%\caption{Architecture of the motion prediction network. A transformer encoder processes each radar point cloud frame into a global feature via a CLS token, a bidirectional LSTM captures temporal context, and a skeleton-aware GCN decoder predicts generalized coordinates and contact labels on the kinematic tree graph.}
%\label{fig:network_model}
%\end{figure}

\begin{table}[b]
\centering
\caption{LOOCV Scale Prediction MAE (\%) Averaged Over Participants. Full per-participant breakdown is in Table~\ref{tab:scale_comparison_full} (Appendix).}
\label{tab:scale_comparison}
\begin{tabular}{lc}
\toprule
Method & Mean $\pm$ Std \\
\midrule
Transformer      & 4.503 $\pm$ 2.073 \\
Ridge            & 3.522 $\pm$ 0.938 \\
MT-Lasso         & \textbf{3.407 $\pm$ 1.282} \\
MT-ElasticNet    & 3.597 $\pm$ 1.001 \\
\bottomrule
\end{tabular}
\end{table}

\subsection{Differentiable Forward Kinematics}
\label{sec:fk}

A fully differentiable forward kinematics (FK) module converts the predicted joint angles $\mathbf{q}_t$ into 3D marker positions. For each body $i$ with parent $p(i)$, the world-frame position and orientation are computed recursively as
\begin{equation}
\mathbf{p}_i = \mathbf{p}_{p(i)} + \mathbf{R}_{p(i)} \left( s_i \cdot \mathbf{o}_i \right), \quad
\mathbf{R}_i = \mathbf{R}_{p(i)} \prod_{k \in \mathcal{H}_i} \mathbf{R}_k(q_k),
\end{equation}
where $\mathbf{o}_i \in \mathbb{R}^{3}$ is the translation offset of body $i$ in its parent's frame, $s_i$ is the subject-specific scale factor, $\mathcal{H}_i$ is the ordered sequence of hinge joints attached to body $i$, and $\mathbf{R}_k(q_k)$ is the rotation matrix for hinge joint $k$ computed via the Rodrigues formula.

\begin{table*}[t]
  \caption{Aggregated LOOCV Results (Mean $\pm$ Std Across All Samples)}
  \label{tab:aggregate}
  \centering
  \resizebox{\textwidth}{!}{%
  \begin{tabular}{l ccc ccc cc ccc}
    \toprule
    & \multicolumn{3}{c}{Pose Estimation} & \multicolumn{3}{c}{Per-Angle MPJPE$\downarrow$ (cm)} & \multicolumn{2}{c}{Bone Std$\downarrow$ (cm)} & \multicolumn{3}{c}{Foot Contact} \\
    \cmidrule(lr){2-4} \cmidrule(lr){5-7} \cmidrule(lr){8-9} \cmidrule(lr){10-12}
    Method
      & MPJPE & PA-MPJPE & MPJAE
      & $0^\circ$ & $45^\circ$ & $90^\circ$
      & Markers & FK
      & Prec. & Rec. & F1 \\
    & (cm) & (cm) & ($^\circ$) & & & & & & & & \\
    \midrule
    mmMesh~\cite{xue2021mmmesh}
      & $7.131 \pm 2.568$ & $4.906 \pm 1.729$ & $8.115 \pm 0.641$
      & $6.920 \pm 2.564$ & $6.874 \pm 2.547$ & $7.589 \pm 2.531$
      & $1.099 \pm 0.311$ & --
      & -- & -- & -- \\
    Proposed
      & $\mathbf{6.456 \pm 1.759}$ & $\mathbf{4.735 \pm 1.268}$ & $\mathbf{8.083 \pm 0.884}$
      & $\mathbf{6.158 \pm 1.540}$ & $\mathbf{6.295 \pm 1.521}$ & $\mathbf{6.905 \pm 2.062}$
      & $\mathbf{0.315 \pm 0.088}$ & $0.000 \pm 0.000$
      & $0.923 \pm 0.019$ & $0.947 \pm 0.009$ & $0.935 \pm 0.009$ \\
    \bottomrule
  \end{tabular}%
  }
\end{table*}

\begin{table}[!b]
  \caption{Per-Subject LOOCV Results (Mean and Inter-Subject Std of Subject-Level Means)}
  \label{tab:per_subject}
  \centering
  \resizebox{\columnwidth}{!}{%
  \begin{tabular}{l cc cc cc c}
    \toprule
    & \multicolumn{2}{c}{MPJPE$\downarrow$ (cm)}
    & \multicolumn{2}{c}{PA-MPJPE$\downarrow$ (cm)}
    & \multicolumn{2}{c}{MPJAE$\downarrow$ ($^\circ$)}
    & Contact \\
    \cmidrule(lr){2-3} \cmidrule(lr){4-5} \cmidrule(lr){6-7} \cmidrule(lr){8-8}
    Subject
      & Proposed & mmMesh
      & Proposed & mmMesh
      & Proposed & mmMesh
      & F1$\uparrow$ \\
    \midrule
    S1  & 7.159 & \textbf{6.434} & 5.297 & \textbf{4.966} & 9.339 & \textbf{8.536} & 0.933 \\
    S2  & 6.817 & \textbf{6.722} & \textbf{5.149} & 5.210 & \textbf{8.418} & 9.119 & 0.935 \\
    S3  & 5.902 & \textbf{5.814} & \textbf{4.426} & 4.561 & 8.655 & \textbf{8.085} & 0.940 \\
    S4  & 6.021 & \textbf{5.468} & \textbf{4.165} & 4.254 & 8.223 & \textbf{7.767} & 0.926 \\
    S6  & \textbf{7.491} & 8.165 & \textbf{5.637} & 5.771 & 8.114 & \textbf{7.882} & 0.923 \\
    S7  & 6.946 & \textbf{6.347} & \textbf{4.744} & 4.783 & 9.200 & \textbf{8.251} & 0.944 \\
    S8  & \textbf{6.552} & 8.086 & \textbf{4.482} & 4.995 & \textbf{7.979} & 8.961 & 0.924 \\
    S9  & \textbf{6.416} & 6.828 & \textbf{4.896} & 5.138 & \textbf{6.954} & 7.971 & 0.928 \\
    S10 & \textbf{6.480} & 10.735 & \textbf{4.640} & 5.196 & \textbf{8.232} & 8.514 & 0.931 \\
    S11 & \textbf{5.940} & 8.102 & \textbf{4.405} & 5.042 & 7.675 & \textbf{7.363} & 0.946 \\
    S12 & \textbf{5.312} & 5.648 & 4.315 & \textbf{4.073} & \textbf{6.126} & 6.817 & 0.951 \\
    \midrule
    Mean & \textbf{6.456} & 7.131 & \textbf{4.735} & 4.906 & \textbf{8.083} & 8.115 & 0.935 \\
    Std  & \textbf{$\pm$0.604} & $\pm$1.476 & \textbf{$\pm$0.437} & $\pm$0.453 & $\pm$0.884 & \textbf{$\pm$0.641} & $\pm$0.009 \\
    \bottomrule
  \end{tabular}%
  }
\end{table}

\subsection{Training Objective}
\label{sec:contact_training}
The total loss is $\mathcal{L} = \mathcal{L}_{\text{data}} + \mathcal{L}_{\text{contact}}$.
The {data-tracking loss} comprises a keypoint position term and a velocity term. The keypoint loss is the mean squared error between predicted and ground-truth marker positions:
\begin{equation}
\mathcal{L}_{\text{kp}} = \frac{1}{M} \sum_{j=1}^{M} \left\| \hat{\mathbf{m}}_j - \mathbf{m}_j^* \right\|^2,
\end{equation}
where $\hat{\mathbf{m}}_j$ and $\mathbf{m}_j^*$ are predicted and ground-truth marker positions and $M$ is the number of markers. The velocity term penalizes finite-difference marker velocity deviations:
\begin{equation}
\mathcal{L}_{\text{vel}} = \mathrm{mean}\!\left(\left\| \Delta\hat{\mathbf{m}}\,f_s - \Delta\mathbf{m}^*\,f_s \right\|^2\right),
\end{equation}
where $\Delta\hat{\mathbf{m}}_{t,j} = \hat{\mathbf{m}}_{t+1,j} - \hat{\mathbf{m}}_{t,j}$ denotes finite differences and $f_s = 15$\,Hz is the frame rate.

The {contact classification loss} supervises foot-ground interaction using binary labels derived from the ground-truth skeleton. A foot body is labeled as in contact when its height $y_{i,t}$ above the estimated ground plane $y_{\text{gnd}}$ and its vertical velocity $\dot{y}_{i,t}$ are below thresholds $\delta_{h,i}$ and $\delta_v$ (see Appendix~\ref{sec:loss_details}):
\begin{equation}
c_{i,t} = \mathbf{1}\!\left[y_{i,t} - y_{\text{gnd}} < \delta_{h,i}\right] \cdot \mathbf{1}\!\left[|\dot{y}_{i,t}| < \delta_v\right].
\label{eq:contact_label}
\end{equation}
The contact loss combines binary cross-entropy $\mathcal{L}_{\text{bce}}$ with velocity and height regularizers weighted by the predicted contact probability $\varsigma(\hat{\mathbf{c}})$:
\begin{multline}
\mathcal{L}_{\text{contact}} = \mathcal{L}_{\text{bce}} + \mathrm{mean}\!\left(\varsigma(\hat{\mathbf{c}}) \cdot \left\| \dot{\mathbf{p}}_{\text{foot}} \right\|^2\right) \\
+ \mathrm{mean}\!\left(\varsigma(\hat{\mathbf{c}}) \cdot \mathrm{ReLU}\!\left(y_{\text{foot}} - y_{\text{gnd}} - \delta\right)\right),
\end{multline}
where $\varsigma(\cdot)$ is the sigmoid function, $\hat{\mathbf{c}}$ are predicted contact logits, $\dot{\mathbf{p}}_{\text{foot}}$ is the 3D foot velocity from FK, $y_{\text{foot}}$ is the vertical foot position, and $\delta = 0.02$\,m is a height tolerance. Loss weights are provided in Appendix~\ref{sec:loss_details}.

%% ====================================================================
\section{RESULTS}
%% ====================================================================

We evaluate using four groups of metrics. Pose accuracy is measured via mean per-joint position error (MPJPE) and Procrustes-aligned MPJPE (PA-MPJPE). Mean per-joint angle error (MPJAE) captures the deviation between predicted and ground-truth joint angles. Bone length consistency is the temporal standard deviation of inter-joint distances (Bone Std), computed both from FK body positions and from virtual marker positions. Foot contact classification is assessed by precision, recall, and F1. We evaluate against mmMesh~\cite{xue2021mmmesh}, a widely adopted unconstrained regression baseline~\cite{chen2022mmbody, yang2024mmbat, fan2024diffusion, yang2025mmdear, zhang2024super} with publicly available code, retrained on the same data.

Subject-specific scaling achieves a mean absolute error (MAE) of $3.4 \pm 1.3\,\%$ using multi-task Lasso regression (Table~\ref{tab:scale_comparison}), personalizing the skeleton sufficiently for accurate forward kinematics; full per-participant results are in the Appendix (Table~\ref{tab:scale_comparison_full}). Table~\ref{tab:per_subject} reports per-subject leave-one-subject-out cross-validation (LOOCV) results and Table~\ref{tab:aggregate} summarizes aggregated metrics. Our proposed method achieves $6.456 \pm 1.759$\,cm MPJPE and $8.083 \pm 0.884^\circ$ MPJAE, outperforming mmMesh ($7.131 \pm 2.568$\,cm). Because the skeletal model defines rigid body segments with fixed lengths, bone length variability computed from FK body positions (Bone Std FK) is exactly $0.000$\,cm by construction, compared to $1.099$\,cm for mmMesh. The marker-based bone std ($0.315$\,cm) is slightly nonzero because virtual markers are placed at learned offsets from body origins, but the underlying skeleton remains rigid. Across participants, the proposed method exhibits $2.4\times$ lower inter-subject standard deviation ($\pm 0.604$\,cm vs.\ $\pm 1.476$\,cm for mmMesh), with mmMesh reaching up to $10.7$\,cm for S10, indicating that the constrained output space generalizes more consistently across body types. Across viewing angles, both models degrade at $90^\circ$ (ours: $6.158$--$6.905$\,cm; mmMesh: $6.874$--$7.589$\,cm), but our method remains lower and more homogeneous at every angle, suggesting robustness to radar aspect angle variation.

\begin{figure}[!t]
\centering
\includegraphics[width=\columnwidth]{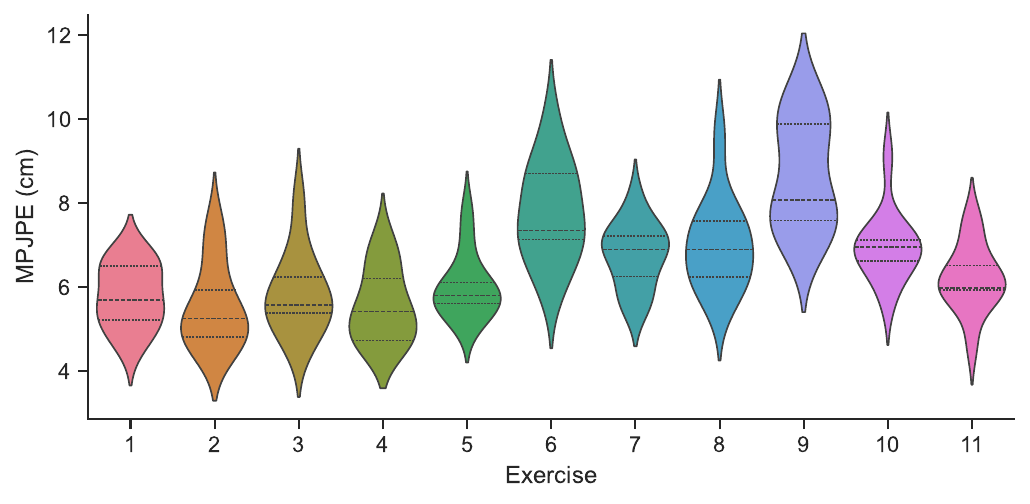}
\caption{Per-exercise MPJPE distribution under LOOCV. Each violin aggregates the per-subject mean MPJPE (averaged across all keypoints) for one exercise, showing inter-subject variability.}
\label{fig:per_exercise}
\end{figure}

\begin{figure}[!t]
\centering
\includegraphics[width=\columnwidth]{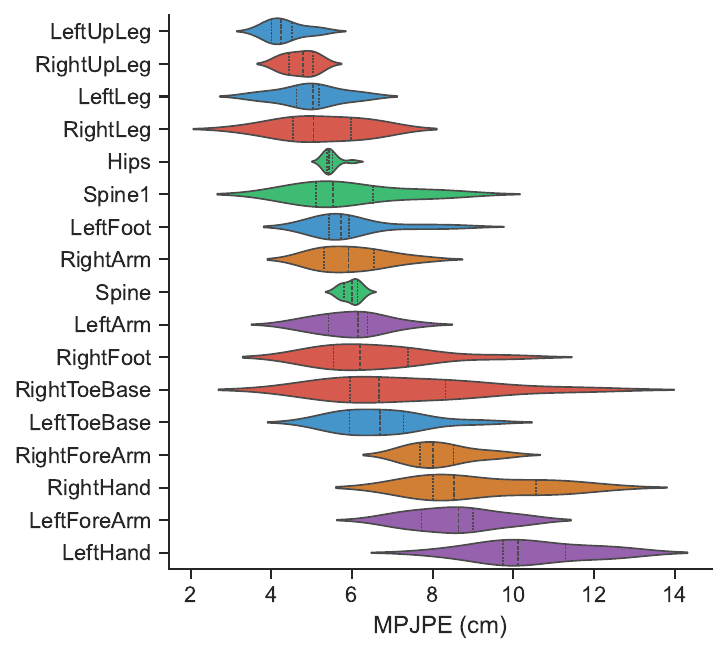}
\caption{Per-keypoint MPJPE distribution under LOOCV. Each violin aggregates the per-subject mean error for one keypoint, revealing which body regions are hardest to reconstruct across participants. The central dashed line marks the median and the outer dashed lines indicate the interquartile range.}
\label{fig:per_keypoint}
\end{figure}

\begin{figure}[!b]
\centering
\includegraphics[width=\columnwidth]{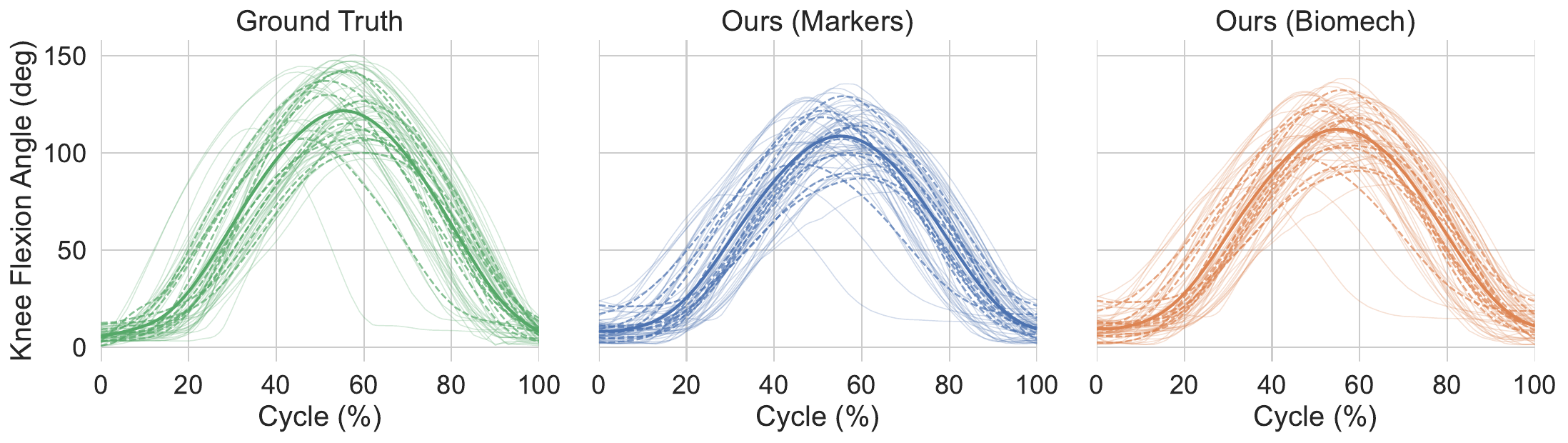}
\caption{Time-normalized knee flexion angles during the squat exercise. Solid lines show the grand mean across all participants, dotted lines show per-participant means, and thin lines show individual squat trajectories.}
\label{fig:knee_angles}
\end{figure}

Fig.~\ref{fig:per_exercise} and Fig.~\ref{fig:per_keypoint} show per-exercise and per-keypoint error distributions. As visible in Fig.~\ref{fig:per_keypoint}, upper-extremity keypoints (hands, wrists, elbows) exhibit higher errors than torso landmarks, a well-documented challenge in radar-based HPE caused by weaker reflections from smaller body surfaces~\cite{sengupta2020mm, sengupta2022mmpose, cao2023task, cao2022joint, hu_mmpose-fk_2024, an2022, an2022fast, su2025posegraphnet, chiang2024enhancing, luo2025, lee2023hupr, ho2024rt}. Fig.~\ref{fig:knee_angles} shows predicted vs.\ ground-truth knee flexion angles for the squat exercise, demonstrating that the model captures the characteristic movement pattern across participants.

\begin{figure}[t]
\centering
\includegraphics[width=\columnwidth]{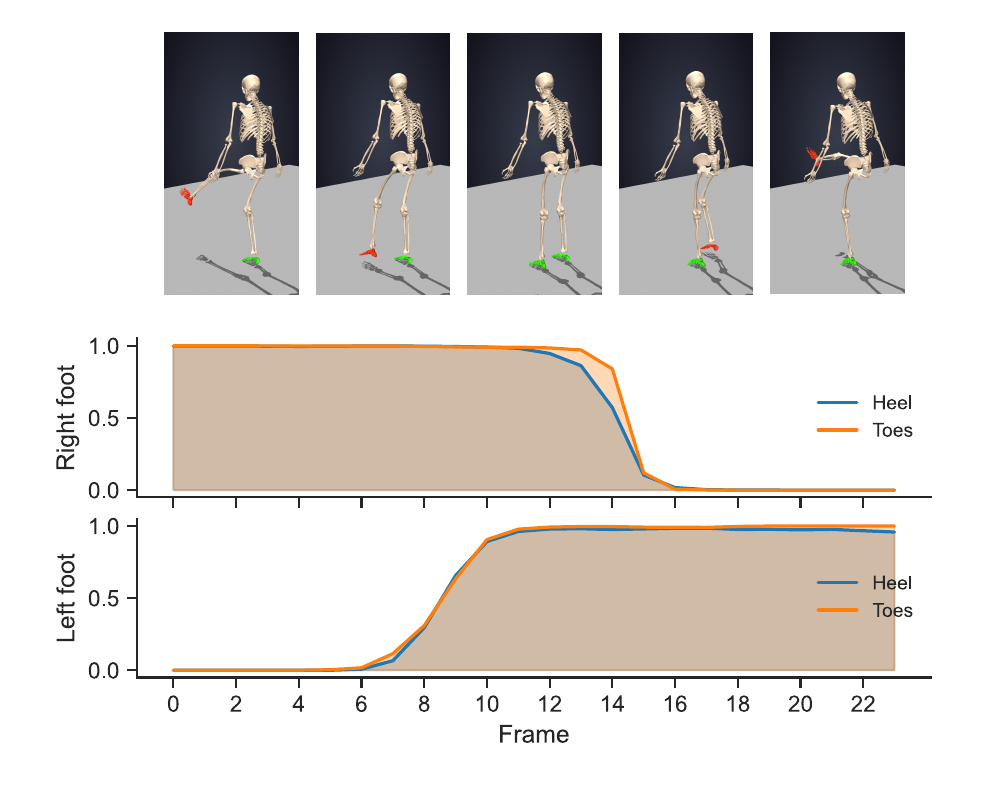}
\caption{Qualitative ground contact classification result. Foot bodies on the biomechanical model are color-coded green when in contact and red when not in contact.}
\label{fig:qualitative}
\end{figure}

Ground contact classification reaches F1 $= 0.935 \pm 0.009$ (precision $0.923$, recall $0.947$), demonstrating reliable stance detection. Fig.~\ref{fig:qualitative} illustrates a qualitative contact classification example.

%% ====================================================================
\section{DISCUSSION}
%% ====================================================================

The results confirm that improving the anatomical fidelity of the human model, rather than solely advancing the learning algorithm, yields substantial benefits for radar-based HPE. Our method outperforms the unconstrained baseline~\cite{xue2021mmmesh} ($6.456$ vs.\ $7.131$\,cm MPJPE) while providing structural guarantees that unconstrained regression cannot offer, namely zero bone length variability and consistent limb proportions. We additionally observe $2.4\times$ lower inter-subject variability and more homogeneous accuracy across viewing angles, consistent with the value of anatomical constraints for radar-based HPE. Mueller et al.\ \cite{mueller2025radproposer} report $6.425$\,cm on the same dataset using raw radar tensors with eight input frames, and Engel et al.\ \cite{engel2025advanced} achieve $6.49$\,cm using point clouds with 50 input frames. Both aggregate multiple frames into a single pose prediction, not directly comparable to our sequence-to-sequence design, yet our error falls within the same range while additionally ensuring kinematic consistency and clinically interpretable joint angles. The elevated error at upper-extremity keypoints reflects a known radar sensing limitation rather than a modeling deficiency~\cite{sengupta2020mm, sengupta2022mmpose, cao2023task, cao2022joint, hu_mmpose-fk_2024, an2022, an2022fast, su2025posegraphnet, chiang2024enhancing, luo2025, lee2023hupr, ho2024rt}, and multi-radar configurations~\cite{lee2023hupr} could address this by increasing point cloud density on smaller body segments, though no such dataset with motion capture ground truth currently exists.
The combination of kinematically consistent joint trajectories and accurate ground contact classification ($\text{F1} = 0.935 \pm 0.009$) provides a basis for downstream biomechanical analysis such as inverse dynamics \cite{werling2023addbiomechanics} to additionally estimate joint torques and ground reaction forces.

The scaled skeletal model is derived entirely from radar without calibration or anthropometric measurements, though direct body measurements can be incorporated for higher fidelity when available \cite{moissenet2017alterations}. This zero-setup property, combined with the privacy-preserving and contactless nature of radar \cite{act2024eu, krauss2024review}, positions the approach for future home deployment for remote monitoring or sensitive medical scenarios where camera-based systems \cite{uhlrich2023opencap} face regulatory barriers or hinder patient compliance, e.g.,\ through stress responses induced by marker-based motion capture \cite{fleischmann2025investigating}.

These results establish a proof of concept that full-body skeletal models can be effectively integrated into radar-based HPE. The consistent accuracy across all 11 healthy participants ($\pm 0.604$\,cm inter-subject std) under LOOCV in a controlled laboratory setting provides a foundation for validation on larger and more diverse cohorts including clinical populations in unconstrained environments, where non-optical ground truth such as IMU-based kinematics estimation \cite{nitschke2024estimating} can replace marker-based motion capture. Future work can further incorporate ground reaction force supervision from force plates to extend the framework from kinematics to full dynamics, and ultimately enable complete contactless and privacy-preserving biomechanical motion reconstruction from radar alone.

\subsection{Scope and Limitations}
\label{sec:limitations}

As a proof-of-concept study, this work establishes feasibility under controlled conditions. The cohort of 11 healthy participants limits conclusions about clinical populations and broader body-type variation, and the reported inter-subject variability is best read as an observation at this sample size. The protocol is nonetheless the most demanding use of the available data, as leave-one-subject-out cross-validation makes every reported result out-of-sample on an unseen participant, and cohorts of this kind remain rare because supervision requires radar and marker-based optical motion capture to be recorded in hardware synchronization. The natural next validation is a rehabilitation clinic or movement laboratory, where the environment remains controlled and the sensor position fixed, so that the principal gap is the population rather than the recording conditions. The bidirectional temporal model spans future as well as past frames, which suits the offline analysis typical of biomechanical assessment; a causal variant would be required for online monitoring.

%% ====================================================================
\section{CONCLUSIONS}
%% ====================================================================

We presented the first integration of a full-body skeletal model into radar-based human pose estimation. Combining differentiable forward kinematics with subject-specific scaling and contact classification, a single FMCW radar recovers biomechanically plausible motion from sparse point clouds without body-worn sensors or cameras, with accuracy comparable to state-of-the-art radar-based methods while additionally ensuring consistent bone lengths and clinically interpretable joint angles. As a proof of concept evaluated on 11 healthy participants in a controlled laboratory setting, these results represent a step toward bringing biomechanical analysis capabilities closer to low-cost, privacy-preserving sensing for rehabilitation monitoring.

\section*{Ethics Statement}
The data used in this study were collected under ethics protocol \#22-437-B, approved by the institutional review board of Friedrich-Alexander-Universitaet Erlangen-Nuernberg. All participants provided written informed consent prior to participation.

\bibliographystyle{IEEEtran}
\bibliography{references}

%% ====================================================================
%% APPENDIX (Supplementary Materials of the journal submission)
%% ====================================================================

\clearpage
\appendix

\renewcommand{\thetable}{S-\Roman{table}}
\renewcommand{\thefigure}{S\arabic{figure}}
\renewcommand{\thesubsection}{S-\Alph{subsection}}
\renewcommand{\thesubsectiondis}{S-\Alph{subsection}.}
\setcounter{table}{0}
\setcounter{figure}{0}
\setcounter{subsection}{0}

\noindent This appendix contains the supplementary materials of the manuscript: virtual marker definitions, details on the scale prediction methods, a complete specification of all loss terms, and network architecture and training hyperparameters.

\subsection{Virtual Marker Definitions}
\label{sec:marker_defs}

\begin{figure}[h]
\centering
\includegraphics[height=7cm]{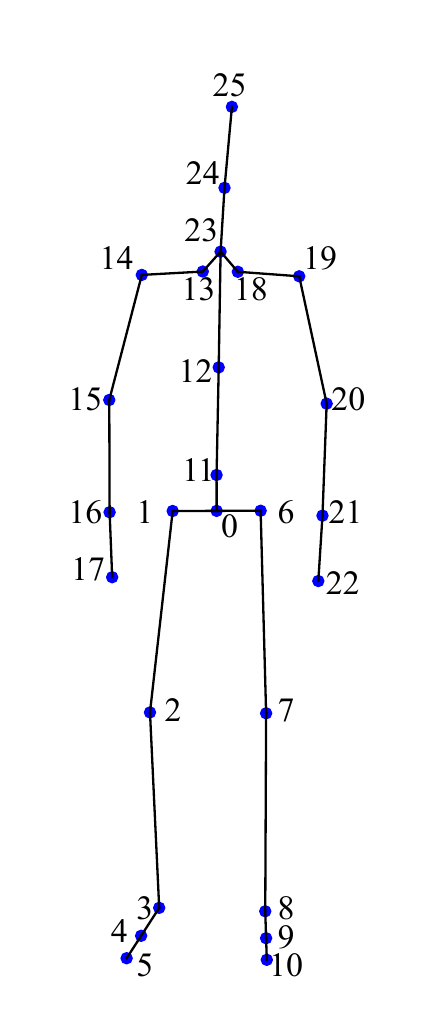}
\caption{The 26-joint ground-truth skeleton from optical motion capture with joint indices. The 17 joints used as virtual markers (indices 0--4, 6--9, 11--12, 14--16, 19--21) are listed in Table~\ref{tab:markers}.}
\label{fig:gt_skeleton}
\end{figure}

The $M = 17$ virtual markers used for supervision are a subset of the 26 OptiTrack keypoints (Fig.~\ref{fig:gt_skeleton}), selected based on proximity to their parent body origin on the biomechanical model ($< 15$\,cm). Table~\ref{tab:markers} lists each marker with its joint index, parent body segment, and anatomical region.

\begin{table}[h]
\centering
\caption{Virtual Marker Definitions}
\label{tab:markers}
\begin{tabular}{clll}
\toprule
Index & Marker & Body Segment & Region \\
\midrule
0  & Hips          & pelvis     & pelvis \\
1  & RightUpLeg    & femur\_r   & thigh \\
2  & RightLeg      & tibia\_r   & shank \\
3  & RightFoot     & calcn\_r   & foot \\
4  & RightToeBase  & toes\_r    & foot \\
6  & LeftUpLeg     & femur\_l   & thigh \\
7  & LeftLeg       & tibia\_l   & shank \\
8  & LeftFoot      & calcn\_l   & foot \\
9  & LeftToeBase   & toes\_l    & foot \\
11 & Spine         & pelvis     & pelvis \\
12 & Spine1        & torso      & trunk \\
14 & RightArm      & humerus\_r & upper arm \\
15 & RightForeArm  & ulna\_r    & forearm \\
16 & RightHand     & hand\_r    & hand \\
19 & LeftArm       & humerus\_l & upper arm \\
20 & LeftForeArm   & ulna\_l    & forearm \\
21 & LeftHand      & hand\_l    & hand \\
\bottomrule
\end{tabular}
\end{table}

\subsection{Scale Prediction from Radar}
\label{sec:scale_prediction}

As a preliminary investigation into whether subject-specific body proportions can be estimated directly from radar data without motion capture, we compare several regression methods that predict the heuristic ground-truth scale factors (Sec.~\ref{sec:scaling}) from radar exercise sequences.

\paragraph{Geometric feature extraction} For each radar frame, we extract 19 handcrafted geometric features: spatial extent (3), centroid (3), standard deviation (3), and interquartile range (3) per axis; horizontal spread within four vertical height bands (4); mean and standard deviation of reflectivity (2); and valid point count (1). These per-frame features are aggregated across time by computing their mean and standard deviation, yielding a 38-dimensional feature vector per participant.

\paragraph{Multi-task Lasso regression} The aggregated geometric features are standardized and fed into a multi-task Lasso model that jointly predicts 10 independent scale residuals, one per unique body segment, with shared $\ell_1$ regularization across outputs to encourage feature selection. As described in Sec.~\ref{sec:scaling}, the residuals are added to a unit baseline and mirrored bilaterally to produce $\mathbf{s} \in \mathbb{R}^{20}$. Additional linear baselines include Ridge regression and multi-task ElasticNet.

\paragraph{Transformer baseline} For comparison, a Transformer-based scaling network shares the same per-frame point cloud encoder architecture as the motion prediction network (Sec.~\ref{sec:motion_net}) but uses a smaller embedding dimension of $d = 64$ and four attention heads. A learnable temporal CLS token aggregates information across frames, and a two-layer MLP with GELU activation predicts the same 10 scale residuals.

All methods are trained with leave-one-subject-out cross-validation to predict the heuristic ground-truth scale factors. Per-participant results are reported in Table~\ref{tab:scale_comparison_full}.

\begin{table}[h]
\centering
\caption{Radar Sensor Parameters}
\label{tab:radar_params}
\begin{tabular}{lc}
\toprule
Parameter & Value \\
\midrule
Frequency & 60\,GHz \\
RF bandwidth & $\approx 1.02$\,GHz \\
Frame rate & 15\,Hz \\
Chirp duration & $\approx 17$\,\textmu s \\
Samples per chirp & 64 \\
ADC sampling frequency & 3.8\,MHz \\
Chirps per frame per TX antenna & 128 \\
TX / RX antennas & 3 / 4 \\
TDM-MIMO virtual channels & 12 \\
Azimuth/Elevation resolution & $\approx 30^\circ$ \\
Field of view & $\pm 60^\circ$ \\
Range resolution & $\approx 14.8$\,cm \\
Unambiguous range & $\approx 9.49$\,m \\
Doppler resolution & $\approx 0.078$\,m\,s$^{-1}$ \\
Unambiguous Doppler velocity & $\approx \pm 5.02$\,m\,s$^{-1}$ \\
\bottomrule
\end{tabular}
\end{table}

\begin{table*}[t]
\centering
\caption{LOOCV Scale Prediction MAE (\%) Per Participant}
\label{tab:scale_comparison_full}
\resizebox{\textwidth}{!}{%
\begin{tabular}{l*{11}{c}c}
\toprule
Method & p1 & p2 & p3 & p4 & p6 & p7 & p8 & p9 & p10 & p11 & p12 & Mean $\pm$ Std \\
\midrule
Transformer      & 2.298 & 2.648 & 4.888 & \textbf{1.919} & 7.433 & 2.653 & 6.047 & \textbf{3.457} & 8.067 & 6.355 & 3.772 & 4.503 $\pm$ 2.073 \\
Ridge            & 3.155 & 3.401 & \textbf{4.165} & 2.075 & 5.428 & \textbf{2.435} & 3.748 & 3.767 & \textbf{4.674} & 3.174 & \textbf{2.719} & 3.522 $\pm$ 0.938 \\
MT-Lasso         & \textbf{1.992} & \textbf{1.529} & 4.416 & 2.690 & \textbf{4.763} & 2.818 & \textbf{3.292} & 3.572 & 6.326 & 2.975 & 3.103 & \textbf{3.407 $\pm$ 1.282} \\
MT-ElasticNet    & 2.906 & 3.058 & 4.242 & 2.718 & 5.236 & 2.717 & 3.364 & 3.562 & 5.787 & \textbf{2.885} & 3.093 & 3.597 $\pm$ 1.001 \\
\bottomrule
\end{tabular}%
}
\end{table*}

\subsection{Loss Function Details}
\label{sec:loss_details}

Tables~\ref{tab:loss_data} and~\ref{tab:loss_contact} list all loss terms and their weights. The individual contact sub-losses are defined below.

\paragraph{Contact BCE loss} Binary cross-entropy between predicted contact logits and heuristic ground-truth labels (Eq.~\eqref{eq:contact_label}):
\begin{equation}
\mathcal{L}_{\text{bce}} = \mathrm{BCE}(\hat{\mathbf{c}}, \mathbf{c}^*),
\end{equation}
where $\hat{\mathbf{c}} \in \mathbb{R}^{T \times 4}$ are the raw logits and $\mathbf{c}^* \in \{0,1\}^{T \times 4}$ are the heuristic labels.

\paragraph{Contact velocity loss} Penalizes foot velocity when contact is predicted:
\begin{equation}
\mathcal{L}_{\text{c,vel}} = \mathrm{mean}\!\left(\varsigma(\hat{\mathbf{c}}) \cdot \left\| \dot{\mathbf{p}}_{\text{foot}} \right\|^2\right),
\end{equation}
where $\varsigma(\cdot)$ denotes the sigmoid function and $\dot{\mathbf{p}}_{\text{foot}}$ is the 3D foot body velocity computed from FK body positions via finite differences.

\paragraph{Contact height loss} Penalizes foot height above the ground plane when contact is predicted:
\begin{equation}
\mathcal{L}_{\text{c,height}} = \mathrm{mean}\!\left(\varsigma(\hat{\mathbf{c}}) \cdot \mathrm{ReLU}\!\left(y_{\text{foot}} - y_{\text{gnd}} - \delta\right)\right),
\end{equation}
where $y_{\text{foot}}$ is the vertical position of the foot body from FK, $y_{\text{gnd}}$ is the ground offset, and $\delta = 0.02$\,m is a tolerance threshold.

\begin{table}[h]
\centering
\caption{Data-Tracking Loss Terms $\mathcal{L}_{\text{data}}$}
\label{tab:loss_data}
\begin{tabular}{llc}
\toprule
{Term} & {Description} & {Weight} \\
\midrule
$\mathcal{L}_{\text{kp}}$ & Marker position MSE (cm$^2$) & 1.0 \\
$\mathcal{L}_{\text{vel}}$ & FD marker velocity MSE (cm/s)$^2$ & 0.5 \\
\bottomrule
\end{tabular}
\end{table}

\begin{table}[h]
\centering
\caption{Contact Classification Loss Terms $\mathcal{L}_{\text{contact}}$}
\label{tab:loss_contact}
\begin{tabular}{llc}
\toprule
{Term} & {Description} & {Weight} \\
\midrule
$\mathcal{L}_{\text{bce}}$ & Contact classification (BCE) & 2.0 \\
$\mathcal{L}_{\text{c,vel}}$ & Foot velocity during contact & 1.0 \\
$\mathcal{L}_{\text{c,height}}$ & Foot height during contact & 1.0 \\
\bottomrule
\end{tabular}
\end{table}

\subsection{Network Architecture Details}
\label{sec:arch_details}

Table~\ref{tab:arch_summary} summarizes all architectural hyperparameters. We describe each component below.

{Point cloud encoder.} Per-point features are projected via a linear layer from $F = 7$ to $d = 256$ dimensions. A learnable CLS token is prepended to the sequence, yielding $N + 1$ tokens. A single transformer block processes the sequence with multi-head self-attention (8 heads, head dimension 32) followed by a two-layer feed-forward network (hidden dimension 1024, GELU activation). Both sub-layers use pre-layer normalization and residual connections with dropout ($p = 0.1$). The CLS token output is projected through a final linear layer to produce the frame representation $\mathbf{h}_t \in \mathbb{R}^{256}$. All $T$ frames are processed in parallel as a single batch.

{Temporal BiLSTM.} A single-layer bidirectional LSTM with hidden size 128 per direction (256 total) processes the sequence of frame features. A linear output projection maps to $d = 256$, followed by layer normalization and a residual connection: $\mathbf{z}_t = \mathrm{LN}(\mathrm{LSTM}(\mathbf{h}_t) + \mathbf{h}_t)$.

{GCN backbone and decoder heads.} The global feature $\mathbf{z}_t$ is projected to $20 \times 128$-dimensional per-body node features via a linear layer. Three residual Chebyshev spectral graph convolution blocks \cite{defferrard2016convolutional} with polynomial order $K = 2$ refine features along the kinematic tree adjacency. Each block consists of two ChebConv layers with ReLU activation and dropout ($p = 0.1$), connected by a residual path. The decoder uses two linear head sets operating on per-body node features.
\begin{itemize}
\item {Joint angle head:} Per-body linear layers predict the corresponding generalized coordinates (37 DOFs total). Left-right body pairs share weights with sign correction masks to enforce bilateral symmetry.
\item {Contact head:} Linear layers on the four foot body nodes (calcaneus and toes, bilateral) produce scalar logits.
\end{itemize}

\begin{table}[h]
\centering
\caption{Network Architecture and Training Hyperparameters}
\label{tab:arch_summary}
\begin{tabular}{llc}
\toprule
{Component} & {Parameter} & {Value} \\
\midrule
\multirow{5}{*}{Transformer} & Embedding dim ($d$) & 256 \\
& Attention heads & 8 \\
& FFN hidden dim & 1024 \\
& Transformer blocks & 1 \\
& Dropout & 0.1 \\
\midrule
\multirow{2}{*}{BiLSTM} & Hidden size / direction & 128 \\
& Layers & 1 \\
\midrule
\multirow{3}{*}{GCN} & Node feature dim & 128 \\
& Residual ChebConv blocks & 3 \\
& Chebyshev order ($K$) & 2 \\
\midrule
\multirow{5}{*}{Training} & Optimizer & AdamW \\
& Learning rate & $10^{-4}$ \\
& Weight decay & $10^{-4}$ \\
& Batch size & 16 \\
& Gradient clip norm & 1.0 \\
& Temporal window ($T$) & 64 \\
\midrule
\multirow{5}{*}{Loss weights} & $w_{\text{kp}}$ & 1.0 \\
& $w_{\text{vel}}$ & 0.5 \\
& $w_{\text{bce}}$ & 2.0 \\
& $w_{\text{c,vel}}$ & 1.0 \\
& $w_{\text{c,height}}$ & 1.0 \\
\bottomrule
\end{tabular}
\end{table}

{Training protocol.} We train with temporal windows of $T = 64$ frames at $f_s = 15$\,Hz using the AdamW optimizer with learning rate $10^{-4}$, weight decay $10^{-4}$, batch size 16, and gradient clipping at norm 1.0. Bilateral augmentation randomly mirrors training sequences with 50\,\% probability to improve left-right symmetry. We evaluate using leave-one-subject-out cross-validation (LOOCV) across all 11 participants, where each participant serves as the test subject while the remaining 10 are split into 9 for training and 1 for validation.

\end{document}